\pdfoutput=1

\documentclass[11pt]{article}

\usepackage[final]{acl}

\usepackage{times}
\usepackage{latexsym}

\usepackage[T1]{fontenc}

\usepackage[utf8]{inputenc}

\usepackage{microtype}

\usepackage{inconsolata}

\usepackage{graphicx}

\usepackage{pifont}
\usepackage{amsthm,amsmath,amssymb}
\usepackage{multirow} 
\usepackage{tcolorbox}
\usepackage{hyperref}
\usepackage{enumitem}
\usepackage{multicol}
\usepackage{booktabs} 
\usepackage{fontawesome}
\usepackage{array}
\usepackage{colortbl} 

\title{Tool Zero: Training Tool-Augmented LLMs via Pure RL from Scratch}

\author{
 \textbf{Yirong Zeng\textsuperscript{1}},
 \textbf{Xiao Ding\thanks{Corresponding author. Email: xding@ir.hit.edu.cn} \textsuperscript{1}},
 \textbf{Yutai Hou\textsuperscript{2}}, 
 \textbf{Yuxian Wang\textsuperscript{2}},
 \textbf{Li Du\textsuperscript{4}},
 \textbf{Juyi Dai\textsuperscript{1}}, \\
 \textbf{Qiuyang Ding\textsuperscript{2}}, 
 \textbf{Duyu Tang\textsuperscript{2}},
 \textbf{Dandan Tu\textsuperscript{2}},
 \textbf{Weiwen Liu\textsuperscript{3}},
 \textbf{Bing Qin\textsuperscript{1}},
 \textbf{Ting Liu\textsuperscript{1}},
\\
 \textsuperscript{1}Harbin Institute of Technology SCIR Lab,
 \textsuperscript{2}Huawei Technologies Co., Ltd, \\
 \textsuperscript{3}Shanghai Jiao Tong University,
 \textsuperscript{4}Beijing Academy of Artificial Intelligence \\
\\
}

\begin{document}
\maketitle
\begin{abstract}
Training tool-augmented LLMs has emerged as a promising approach to enhancing language models' capabilities for complex tasks.
The current supervised fine-tuning paradigm relies on constructing extensive domain-specific datasets to train models. 
However, this approach often struggles to generalize effectively to unfamiliar or intricate tool-use scenarios. 
Recently, reinforcement learning (RL) paradigm can endow LLMs with superior reasoning and generalization abilities.
In this work, we address a key question: Can the pure RL be used to effectively elicit a model's intrinsic reasoning capabilities and enhance the tool-agnostic generalization?
We propose a dynamic generalization-guided reward design for rule-based RL, which progressively shifts rewards from exploratory to exploitative tool-use patterns.
Based on this design, we introduce the \textit{Tool-Zero} series models. 
These models are trained to enable LLMs to autonomously utilize general tools by directly scaling up RL from \textit{Zero} models (i.e., base models without post-training).
Experimental results demonstrate that our models achieve over 7\% performance improvement compared to both SFT and RL-with-SFT models under the same experimental settings. 
These gains are consistently replicated across cross-dataset and intra-dataset evaluations, validating the effectiveness and robustness of our methods. 

\end{abstract}

\section{Introduction}
Integrating LLMs with external tools has emerged as a pivotal advancement, significantly enhances their ability to address complex tasks \citep{qu2025tool,wang2024executable}. 
It opens up many practical uses across different fields. 
For example, it supports the automation of reasoning tasks \citep{jin2025search,manduzio2024improving}, and enables Agent applications \citep{gunter2024apple,chen2024octopus}.
A tool-augmented model can respond to a user's query by invoking and executing external tools.
In this paper, tools are used interchangeably with APIs, functions, and plugins.

\begin{figure}[t]
    \centering
  \includegraphics[width=0.99\linewidth]{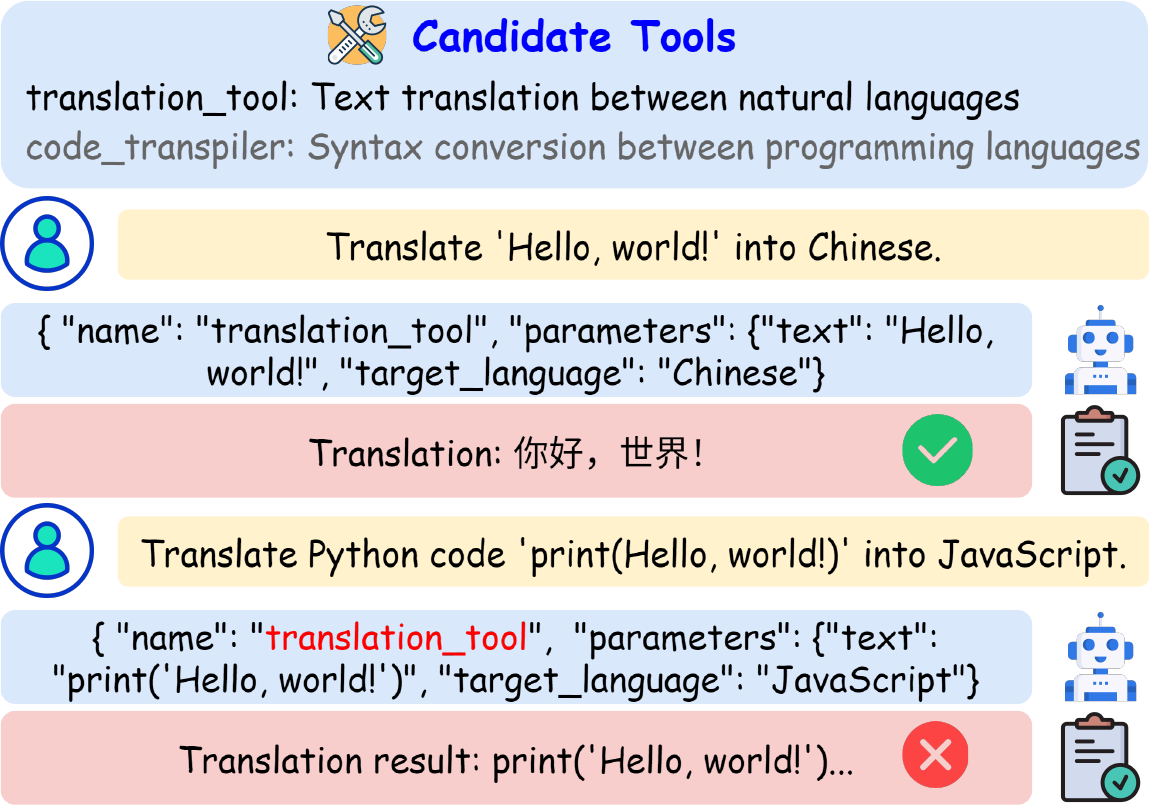}
  \caption{A response demonstration of a tool-augmented model trained in SFT paradigm.
    The model fails to recognize similar but unfamiliar task contexts (e.g., code transpilation), highlighting limited generalization to unseen tool-use scenarios.
    }
  \label{fig:motivation}
\end{figure}

Current approaches to enhance tool-use capability involve synthesizing extensive tool-use trajectories with advanced language models, followed by SFT on the generated data \citep{liu2024toolace,lin2024hammer,chen2025acebench}. 
Under this paradigm, models achieves satisfactory performance in the scenarios that have the same distribution as the training data \citep{bfclv3}. 
However, these SFT-trained models primarily engage in imitating surface-level patterns rather than internalizing the reasoning process, tend to memorize the training trajectories rather than developing robust, intrinsic reasoning capabilities \citep{chen2025acebench}.
Consequently, they exhibit limited generalization ability when applied to unseen scenarios, as elaborated in preliminary study in Section \ref{sec:motivation}.

An illustrative example in Figure \ref{fig:motivation} demonstrates that while the model correctly addresses a natural language translation task, it fails to appropriately invoke tools for code transpilation. 
Specifically, the model interprets "translation" solely at the natural language level, failing to recognize the code transpilation scenario implicit in the user’s query. 
This mismatch underscores a critical limitation: current models lack the intrinsic reasoning abilities to discern nuanced task contexts. 
Enabling LLMs with genuine reasoning capabilities to overcome such generalization barriers has thus become an urgent research imperative.

Recent studies have demonstrated that simple rule-based R1-style RL \citep{deepseekai2025r1}, even without SFT, can significantly enhance LLMs’ complex reasoning capabilities \citep{zeng2025simplerl,lu2025ui,shen2025vlm}. 
This paradigm inspires us to extend pure RL to the tool learning domain, aiming to address the generalization limitations by eliciting models’ intrinsic reasoning abilities. 
To this end, we propose a dynamic generalization-guided reward design for rule-based RL. 
This approach employs a progressive reward strategy: it first promotes early-stage exploratory behavior to cultivate intrinsic reasoning, then refines these capabilities into tool-use patterns focused on final-task precision. 
This design effectively resolves the exploration-exploitation dilemma in open-domain tool learning, bridging the gap between reasoning generalization and task-specific tool use.

To evaluate generalization, we conducted extensive experiments across diverse function-calling benchmarks. Results demonstrate that our proposed Tool-Zero 7B/32B models, trained using our method, significantly outperform both SFT models and RL-with-SFT baselines. 
For example, Tool-Zero-7B achieves a 7.14\% performance improvement compared to SFT model ToolACE-8B. 
Notably, it also surpasses the RL-with-SFT model ToolRL-7B by 7.18\%. 
Additionally, these gains are consistently replicated in both cross-dataset and intra-dataset evaluations.


\section{Related Work}
To contextualize our approach, we survey prior research on tool learning and its integration with large language models.
\subsection{Tool Learning}
Enhancing LLMs with external tools has emerged as a pivotal direction for addressing complex tasks in open domains \citep{qu2025tool, wang2024executable}. Typical applications include integrating LLMs with search engines \citep{zhang2024ecoact, lazaridou2022internet, shuster2022blenderbot}, calculators \citep{nakano2021webgpt}, and Python interpreters \citep{wang2024executable, song2024adaptive, chen2022program}. A dominant paradigm for equipping LLMs with external tools is imitation learning, where language models are trained via imitation on human-labeled datasets. This framework typically involves constructing large-scale supervised tool-use datasets \citep{prabhakar2025apigen, liu2024apigen, liu2024toolace} and applying either SFT \citep{zhang2024xlam, zhang2024ecoact, toolllm} or direct preference optimization (DPO) reinforcement learning \citep{zeng2025boosting, yu2024steptool}, enabling models to autonomously create and invoke tools. However, this paradigm faces challenges in enabling LLMs to generalize across diverse tools with varied argument structures and domains, highlighting a critical gap in tool-agnostic generalization.

\subsection{Tool-Integrated Reasoning with Reinforcement Learning}
RL has gained traction as a more scalable and generalizable training paradigm. 
Models like R1-Zero leverage group relative policy optimization (GRPO) \citep{shao2024deepseekmath} to unlock the model’s reasoning capabilities at test time \citep{deepseekai2025r1, yu2025dapo}. 
This R1-style reasoning paradigm, marking a shift from train-time scaling to test-time scaling \citep{muennighoff2025s1, xia2025generative}, has demonstrated success in mathematics \citep{shao2024deepseekmath}, coding \citep{pan2025metaspatial}.

Recent studies \citep{jin2025search, qian2025toolrl} have explored unlocking tool-integrated reasoning for LLMs, with works like Torl \citep{li2025torl} and ReTool \citep{feng2025retool} achieving promising performance in mathematical tasks by integrating code tools. 
However, their training follows the SFT-then-RL paradigm and remains constrained to single-type tool-use scenarios.
In contrast, our work aims to unlock the model’s tool-agnostic (general-purpose tools) generalization capabilities via pure reinforcement learning scaled directly from \textit{Zero} model.



\section{Problem Statement and Analysis}
\label{sec:motivation}

\textbf{Problem Formulation.}
We first provide the problem formulation of reasoning in tool augmented models.  
It formalizes the integration of external tools into the inference process to solve complex tasks. 
Given a tool set $\mathcal{T} = \{t_1, t_2, \dots, t_n\}$ and a user query $q$, the reasoning trajectory up to step $k$ is defined as:
\begin{equation}
    \small
    \tau_k = \left[ a_1(c_1), o_1 \right], \left[ a_2(c_2), o_2 \right], \dots, \left[ a_k(c_k), o_k \right],
\end{equation}
here, $a_i$ denotes the model's reasoning action (natural language thought) at step $i$, $c_i \subseteq \mathcal{T}$ represents the subset of tools called at step $i$, and $o_i$ denotes the observations received after tool execution, including environment and user feedback.

The model's policy is defined as $\pi: \tau_k \rightarrow a_{k+1}(c_{k+1})$. At each step $k+1$, the model must generate the next reasoning action $a_{k+1}$, select a tool subset $c_{k+1} \subseteq \mathcal{T}$, and formulate parameterized tool invocations for $c_{k+1}$. 
The goal is to enable LLMs with a generalized policy $\pi$ that effectively addresses user queries by producing a sequence of action-observation pairs $(a_t, o_t)$.

\subsection{Preliminary Study}
\label{sec:pre_study}
This section aims to show the generalization challenges faced by tool-augmented models trained in the SFT paradigm and presents the motivation of this paper.
To this end, we conducted the following two preliminary studies:

\begin{figure*}[t]
  \centering
  \includegraphics[width=0.85\linewidth]{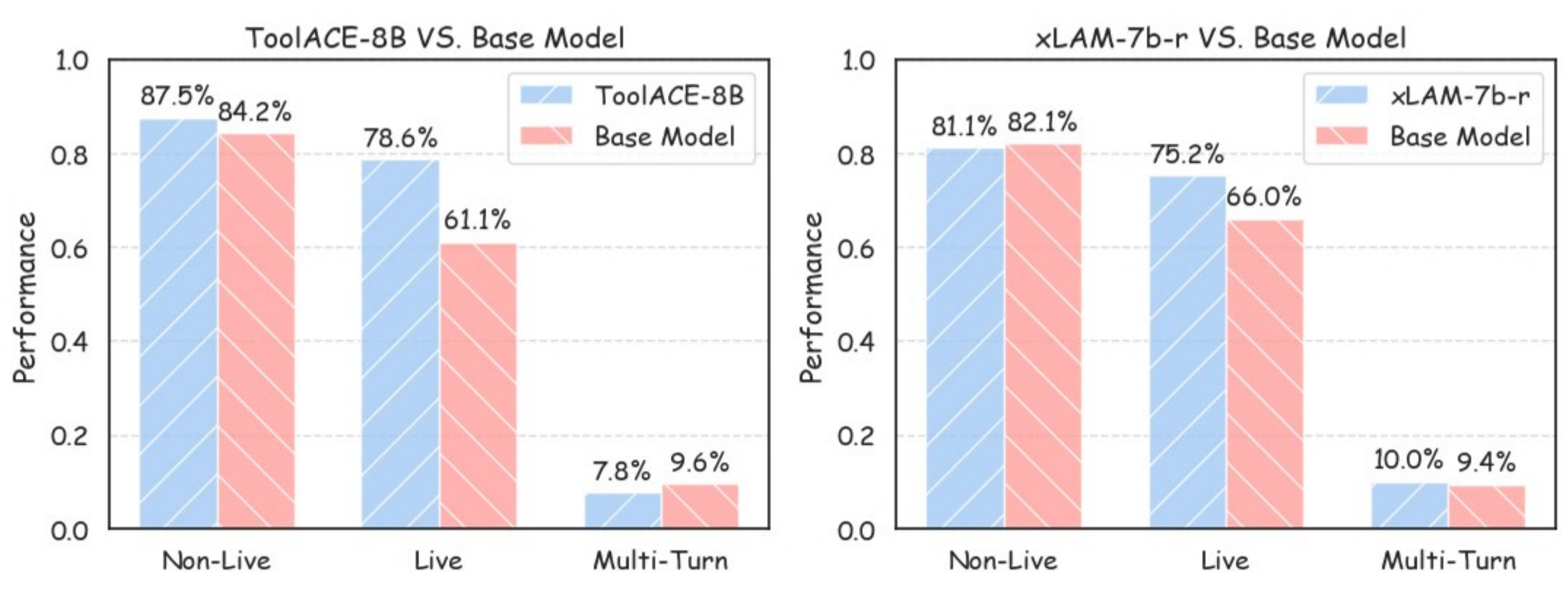}
  \caption{ Intra-Dataset Performance. 
    The improvement on metric (Live) with training-distributed data is significantly greater than that on other metrics.
  SFT struggles with out-of-distribution generalization in open-domain settings.  }
  \label{fig:bfcl_live}
\end{figure*}

\begin{table*}[th]
    \small
    \centering
    \begin{tabular}{l|ccccc|c}
    \toprule
    Models & BFCL-v3 & API-Bank & SealTool & Tool-Alpaca & Nexus Raven & Avg. \\ \midrule
    \ding{168} Granite-20B-FunctionCalling  
      & \ding{174}49.31 & \ding{173}68.53 & \ding{172}92.74 & \ding{173}58.03 & \ding{172}75.15  & \ding{172}\textbf{68.75} \\ 
    \ding{168} Gorilla-OpenFunctions-v2-7B  
      & \ding{173}52.10 & \ding{174}62.50 & \ding{173}91.12 & \ding{174}51.30 & \ding{173}68.46  & \ding{173}65.09\\ 
    \ding{168} xLAM-7B-fc  
      & \ding{172}\textbf{54.75} & \ding{172}72.45 & \ding{174}76.90 & \ding{172}59.00 & \ding{174}57.50  & \ding{174}64.12 \\ 
    
    \midrule 
    
    \ding{171} Llama-3.1-8B-Instruct  
    & \ding{174}50.87 & \ding{174}69.92 & \ding{174}89.27 & \ding{174}59.36 & \ding{174}67.30  & \ding{174}67.34 \\ 
    \ding{171} Qwen2.5-7B-Instruct  
      & \ding{173}53.69 & \ding{172}70.76 & \ding{173}91.07 &\ding{173}60.24& \ding{173}72.24   & \ding{173}69.64 \\ 
    \ding{171} GPT-3.5-Turbo-0125  
    & \ding{172}\textbf{53.91} & \ding{173}70.71 & \ding{172}93.51 & \ding{172}62.50 & \ding{172}82.86  & \ding{172}\textbf{72.62} \\ 
    
    \bottomrule
    \end{tabular}
    \caption{Cross-Dataset Performance of SFT models (\ding{168}) and foundation models (\ding{171}). 
    Smaller ranking numbers (circled numbers) in each column indicate larger values.
    Inconsistent performance of SFT models across benchmarks, indicating
    SFT enhances in-distribution performance but weakens generalization to unseen scenarios.}
    \label{tab:eval_data}
\end{table*}

\noindent
(1) \textbf{Intra-Dataset Performance}. 
We compared two SFT-trained models: \textit{ToolACE-8B} (Llama3.1-8b-inst finetuned on ToolACE \citep{liu2024toolace}) and \textit{xLAM-7B-r} (Mistral-7b finetuned on xLAM \citep{liu2024apigen}), evaluated on the BECL benchmark \citep{bfclv3} (comprising \textit{Single-turn (Non-Live, Live)} and \textit{Multi-turn} subsets).
Notably, both training datasets use LLM-synthesized data to mimic real-world scenarios, with distributions aligned to BFCL-Live (details in Section \ref{sec:data}).
Results in Figure \ref{fig:bfcl_live} show significant improvements on the Live metric (reflecting in-distribution performance), such as a improvement from 61.1 to 78.6.
Conversely, exhibiting negligible gains or regressions on Non-Live and Multi-Turn subsets (e.g., 9.6 \(\rightarrow\) 7.8).
This suggests that SFT struggles with out-of-distribution generalization in open-domain settings.
For instance, single-turn training data fails to transfer to multi-turn scenarios, and simple tool use patterns do not generalize to complex, interdependent tool chains.

\noindent
(2) \textbf{Cross-Dataset Performance}. 
We extended our evaluation to diverse benchmarks (details in Appendix \ref{sec:benchmark}), inspired by \citealp{lin2024hammer}.
Notably, these benchmarks encompass varied tool-use scenarios(e.g., candidate tools, contextual domains, and invocation formats (JSON vs. Python code)).
Results in Table \ref{tab:eval_data} reveal inconsistent performance of existing tool-use models across benchmarks.
For example, while xLAM-7B-fc achieved top performance on BFCL, it suffered significant degradation on two others, leading to the lowest overall average score. 
In contrast, the foundation models demonstrated more consistent cross-dataset performance.
Therefore, this result highlighting a critical issue: SFT enhances in-distribution performance but weakens generalization to unseen scenarios. (e.g., novel tools, invocation formats).
  
In summary, our analysis reveals a fundamental trade-off in the SFT paradigm: while it enhances in-distribution tool-use accuracy, it severely limits generalization to unseen scenarios.
To address this, we propose adopting a pure RL framework for tool learning, designed to dynamically balance exploration of new tool interactions with exploitation of task-relevant patterns.

\begin{figure*}[t]
    \centering
  \includegraphics[width=0.95\linewidth]{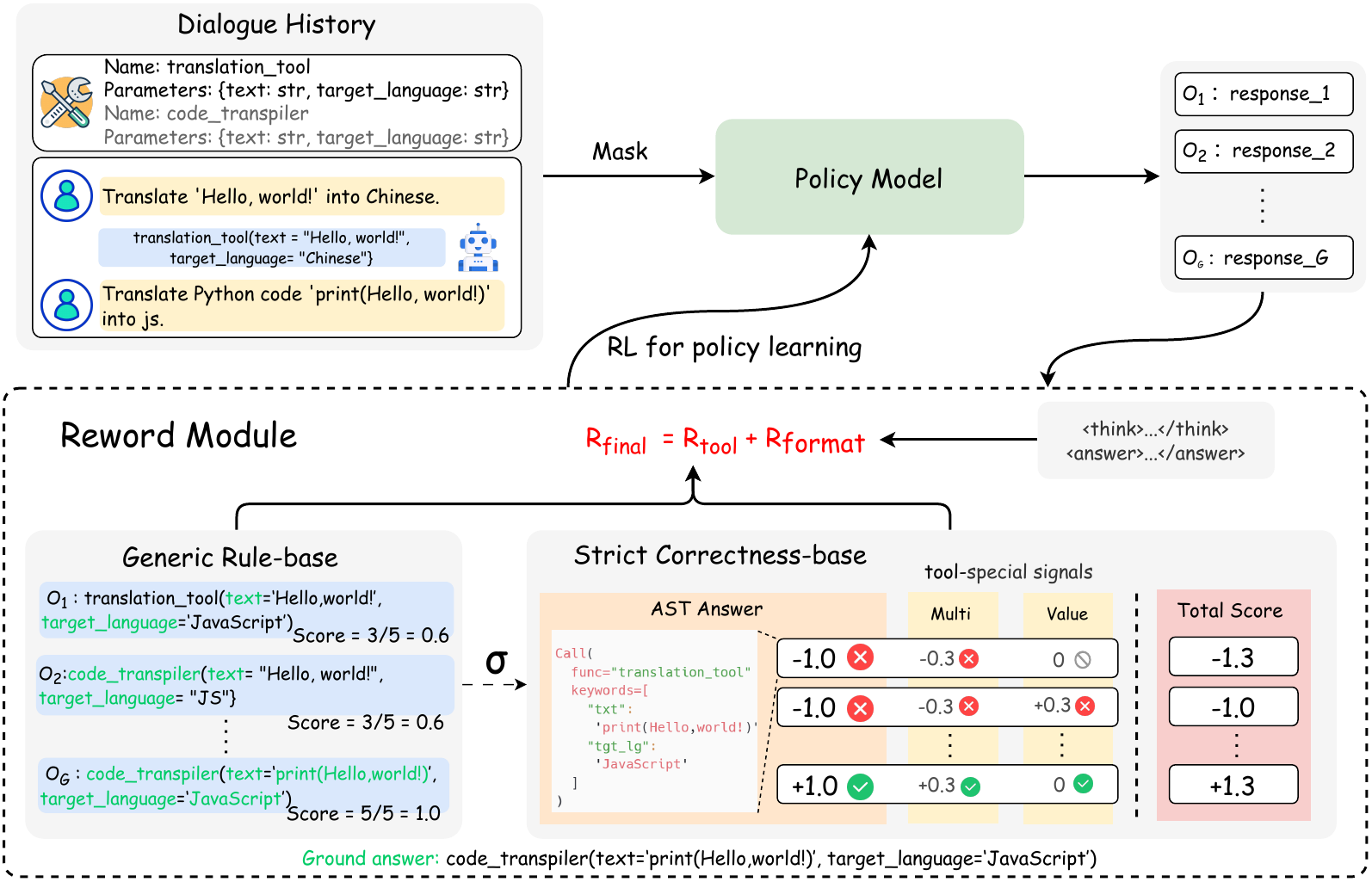}
  \caption{
    The overall architecture of \textit{GG-GRPO} introduces a dynamic generalization-guided reward design for rule-based RL. 
    It progressively shifts the reward mechanism from a fine-grained generic reward to a strict answer correctness reward.
  }
  \label{fig:main}
\end{figure*}

\section{Method}
In this section, we provide a detailed introduction to our method.
Figure \ref{fig:main} shows the overall architecture of our proposed dynamic \underline{G}eneralization-\underline{G}uided reward strategy for \underline{GRPO} (GG-GRPO).

\subsection{Training Data Preparation}
\label{sec:data}
The following data are utilized in RL training.
(1) ToolACE \citep{liu2024toolace}:  It is a general-purpose tool-use dataset, where the model learns when to invoke tools and when to respond directly, thereby enhancing decision-making in multi-step interactions.
(2) xLAM \citep{liu2024apigen}: This is a compositional dataset that requires one or multiple tool calls per turn. It encourages the model to reason about tool dependencies and actively plan diverse tool-calling actions.
We also include an irrelevance-augmented subset\footnote{https://huggingface.co/datasets/MadeAgents/xlam-irrelevance-7.5k} originating from xLAM.

\textbf{Data Filtering}. Since these datasets are generated by potentially unstable LLMs, they often contain non-standard formats for Abstract Syntax Tree (AST) parsing and GRPO training.
We standardize the data by filtering out samples with invalid (1) tool calls (i.e., can be parsed in JSON or Python code format), (2) candidate tools (can be parsed in JSON format).

\textbf{Multi-Turn Augment}. Due to the lack of multi-turn tool-calling trajectories in xLAM, we have augmented it. 
The following four strategies were employed:
(1) single-turn combination: concatenate related single-turn dialogs into multi-turn sequences.
(2) tool removal: randomly remove one tool and reintroduce it in subsequent turns.
(3) parameter clarification: randomly mask a parameter value to prompt user clarification.
(4) result validation: randomly remove tools, delete parameters, or alter values in ground-truth answer to simulate user challenge the response.
We present the statistical information of the proceeding data in Table \ref{tab:data}.
\begin{table}[th]
    \small
    \centering
    \setlength{\tabcolsep}{4pt}
    \begin{tabular}{ccc|cc}
        \toprule
        & \multicolumn{2}{c}{xLAM} & \multicolumn{2}{c}{ToolACE} \\ \midrule
        & Single-T. & Multi-T. & Single-T. & Multi-T. \\ \hline
        Raw Data & 67500 & 0 & 102154 & 2000 \\
        Multi-Aug. & 0 & 10254 & 0 & 0 \\ \hline
        After & 67500 & 10254 & 97300 & 1966 \\ \bottomrule
    \end{tabular}
    \caption{Data statistics for xLAM and ToolACE in data processing.}
    \label{tab:data}
\end{table}

\textbf{Data Mask}. 
Recent studies \citep{lin2024hammer, chen2025acebench} have shown that naming preferences in tool descriptionss can significantly degrade model robustness when testing environments diverge from training conventions. 
To mitigate this issue, we adopt a name-masking strategy aligned with \citealp{lin2024hammer}, which masks function names (e.g., \texttt{calculate\_sum} → \texttt{func\_1}) and parameter names (e.g., \texttt{input\_list} → \texttt{param\_1}).
It redirects the model’s attention to tool descriptions and argument semantics, reduces overfitting to superficial naming patterns, thereby improving tool-agnostic reasoning in open-domain settings.

\subsection{Generalization Guided Reward Design}  
To enhance tool generalization within the GRPO framework, we introduce a \textit{dynamic generalization-guided reward design} that combines flexible exploration with structured convergence. 
Building on prior rule-based reward mechanisms \citep{qian2025toolrl, li2025torl, jin2025search}, our formulation decomposes the total reward $\mathcal{R}_{\text{final}}$ into two components:  
\begin{equation}
\mathcal{R}_{\text{final}} = \mathcal{R}_{\text{format}} + \mathcal{R}_{\text{tool}},
\end{equation} 
where $\mathcal{R}_{\text{format}}$ enforces tool invocation format correctness, and $\mathcal{R}_{\text{tool}}$ drives generalization through a \textit{progressive reward strategy}.  

Our strategy balances initial model exploration with final task precision via two stages:  

\noindent
\textbf{(1) General Rule-Based Reward.}  
During early training iterations, we use a lenient, fine-grained reward to elicit the model’s inherent generalization capabilities. 
This reward ($r_{\text{general}}$) measures the semantic overlap (e.g., tool name, argument name and value) between the model’s response $y$ and the ground-truth answer $y^\ast$ by splitting both into tokenized elements using delimiters, e.g., \texttt{()[],.: '"=}, forming sets $\mathcal{Y} = \{\text{tokens}(y)\}$ and $\mathcal{Y}^\ast = \{\text{tokens}(y^\ast)\}$. The overlap rate is then calculated as:  
\begin{equation}
r_{\text{general}} = -0.5 + \frac{|\mathcal{Y} \cap \mathcal{Y}^\ast|}{|\mathcal{Y}^\ast|} \in [-0.5, +0.5],
\end{equation}
This allows the model to receive partial credit for incomplete but semantically relevant responses, encouraging broad exploration of tool-use patterns during low-capability phases.  

\noindent
\textbf{(2) Strict AST-Based Reward.}
As training progresses, we transition to a strict \emph{Abstract Syntax Tree (AST)-based check} to enforce task-specific tool integration. This stage verifies the structural and semantic correctness of tool invocations (e.g., API argument validity, multi-tool dependency chains) by comparing the generated tool call syntax against a reference AST $\mathcal{T}^\ast$:  
\begin{equation}
    \small
    r_{\text{ast}} = 
        \begin{cases} 
            1 & \text{if } \text{AST}(y) \equiv \mathcal{T}^\ast, \\
            0 & \text{otherwise}.
        \end{cases}
\end{equation}
This ensures the model converges to precise, tool-agnostic reasoning that adheres to complex tool specifications.  

To further enhance the model's tool-integrated reasoning in context, we incorporate two \textit{tool-specific feedback} signals into the strict reward $\mathcal{R}_{\text{strict}}$:
(1) {\textit{multi}-tool collaboration}: $+0.3$ reward for reinforcing correct collaborative tool usage patterns in multi-step tool chains or dialogs.
(2) {parameter \textit{value} error}: $+0.3$ penalty per invalid parameter value to enforce exploring precise context grounding because it demands high-order reasoning from context \citep{zeng2025boosting,lin2024hammer}.

\textbf{Switching Trick}. The dynamic shift from a general reward $r_{\text{general}}$ to a strict reward $r_{\text{strict}}$ is governed by a sigmoid-based decay function:  
\begin{equation}
\small
\mathcal{R}_{\text{tool}} = \sigma(t, m) \cdot r_{\text{strict}} + \left(1 - \sigma(t, m)\right) \cdot r_{\text{general}},
\end{equation}  
where $\sigma(t, m) = \frac{1}{1 + e^{-\kappa(t - m)}}$ is a sigmoid function with steepness $\kappa$,
$t$ and ${m}$ are the current training step and transition midpoint, respectively.
This automated transition trick can avoid abrupt reward changes that may destabilize training.

\textbf{Format Reward.} The format reward $\mathcal{R}_{\text{format}} \in \{0, 1\}$ checks whether the model output contains all required special tokens in the correct order (i.e., \texttt{<think>...</think><answer>...</answer>}).

Overall, our strategy first nurtures broad generalization capabilities and then refines them into structured tool-use behaviors, effectively addressing the exploration-exploitation dilemma in open-domain tool learning.

\subsection{RL Training with Generalization-guided Reward}
\label{sec:method}
To train LLMs from \textit{Zero} model through scaling reinforcement learning, we employ Group Relative Policy Optimization (GRPO) \citep{deepseekai2025r1, shao2024deepseekmath}, that unlocks the model’s reasoning capabilities at test time.

GRPO foregoes the critic model and estimates the baseline from group scores instead.
For each question $q$, GRPO generates $G$ completions $\{o_1, o_2, \dots, o_G\}$ using $\pi_{\theta_{\text{old}}}$, then optimizes $\pi_{\theta}$ by maximizing the following objective:
\begin{equation}
    \small
    \begin{aligned}
    \mathcal{J}_{\text{GRPO}}(\theta) =& \mathbb{E}_{q \sim P(Q), \{o_i\} \sim \pi_{\theta_{\text{old}}}} \Bigg\{ 
    \frac{1}{G} \sum_{i=1}^{G} \frac{1}{|o_i|} \sum_{t=1}^{|o_i|} \\
    & \min \left[ \rho_{i,t} A_i, \text{clip}(\rho_{i,t}, 1\!-\!\epsilon, 1\!+\!\epsilon) A_i \right] 
    \Bigg\}
    \end{aligned}
\end{equation}
where the importance ratio $\rho_{i,t}$ is defined as:
\begin{equation}
\rho_{i,t} \triangleq \frac{\pi_{\theta}(o_{i,t} | q, o_{i,<t})}{\pi_{\theta_{\text{old}}}(o_{i,t} | q, o_{i,<t})},
\end{equation}
here, $\epsilon$ is a hyperparameter. 
Following \citealp{yu2025dapo}, we remove the KL divergence regularization from the GRPO objective.
And $A_i$ is the advantage computed using a group of rewards $\{r_1, r_2, \dots, r_G\}$ corresponding to the completions within each group:
\begin{equation}
A_i = \frac{r_i - \text{mean}(\{r_1, r_2, \dots, r_G\})}{\text{std}(\{r_1, r_2, \dots, r_G\})}.
\end{equation}

In the reward design of GRPO, we replace the rule-based accuracy reward function with a generalization-guided reward (GG-GRPO) for more effective and adaptive reward computation.
The new reward formula is expressed as follows:
\begin{equation}
r_i = \mathcal{R}_{format}(o_i) + \mathcal{R}_{tool}(o_i).
\end{equation}
where, $\mathcal{R}_{format}(o_i)$ denotes format reward of $o_i$ response, $\mathcal{R}_{{tool}}(o_i)$ denotes dynamic generalization guided reward.

\begin{table*}[th]
    \centering
    \begin{tabular}{l|l|ccc|c}
        \toprule
        Type & Model & Non-Live & Live  & Multi-Turn & \textbf{Overall Acc} \\  \midrule

        \multirow{3}{*}{\ding{168}Vanilla}  
            & Llama-3.1-8B-Instruct & 84.21 & 61.08 & 9.62 & 50.87 \\
            &  Qwen2.5-7B-Instruct & 86.46 & 67.44 & 7.62 & 53.69 \\ 
            & Qwen2.5-32B-Instruct & 85.81 & 74.23 & 17.75 & 59.67 \\

        \midrule
        \multirow{3}{*}{\ding{169}SFT}  
            & Hammer2.1-7b & \underline{88.65} & 75.11 & 23.50 & 61.83 \\
            & ToolACE-8B & 87.54 & 78.59 & 7.75 & 58.42 \\
            & xLAM-7b-r & 81.06 & 75.22 & 10.00 & 54.75 \\

        \midrule
        \multirow{5}{*}{\ding{170}API-based}  
            & GPT-3.5-Turbo-0125 & 83.94 & 64.02 & 19.50 & 53.91 \\
            & GPT-4o-mini-2024-07-18 & 85.21 & 74.41 & 34.12 & 64.10 \\ 
            & GPT-4o-2024-11-20 & {88.10} & \underline{79.83} & \underline{47.62} & \underline{72.08} \\
            & Gemini-2.0-Flash-001 & 84.90 & 79.12 & 17.88 & 60.42 \\
            & Gemini-2.0-Pro-Exp-02-05 & 83.94 & 78.50 & 20.75 & 61.55 \\

        \midrule
        \multirow{7}{*}{\ding{171}R1-like}  & DeepSeek-R1 & 87.35 & 74.41 & 12.38 & 56.89 \\
            &  QwQ-32B & 86.48 & 75.48 & 2.12 & 53.93 \\ 
            &  Tool-N1-7B\textsuperscript{*} & 89.25 & 80.38 &  -  & - \\ 
            &  Tool-N1-14B\textsuperscript{*} & 90.52 & 81.42 &  -  & - \\ 
            &  ToolRL-7B & 82.21 & 74.90 & 18.12  & 58.38 \\ 
        \rowcolor{green!20} & {Tool-Zero-7B } & 88.98 & 80.76 & 25.93 & 65.22 \\  
        \rowcolor{green!20} & {Tool-Zero-32B } & \textbf{90.76} & \textbf{82.43} & \textbf{28.18} & \textbf{67.12} \\ 
        \bottomrule
    \end{tabular}
    \caption{ Comparison on the BFCL-v3.  \textit{Overall Acc} denotes the average performance on three subsets.
        \textsuperscript{*} indicates single-turn tool use models, and multi-turn results are not reported.
        \textbf{Bold} for best performance in R1-like models and \underline{underline} for best performance in the other types.
    }
    \label{tab:overall_res}
\end{table*}

\section{Experiments}
In this section, we show the superiority of our method in performance and robustness across various benchmarks, and in-depth analysis to verify the effectiveness of our method.
\subsection{Experimental Setup}
\label{exp_setup}
In the experiment, we employ the Qwen2.5-7B Base and Qwen2.5-32B Base as \textit{Zero} model, and train with GG-GRPO to get our Tool-Zero-7B and Tool-Zero-32B respectively\footnote{trained with ToolACE dataset in the main experiment}. 
More details in Appendix \ref{sec:exp_setup}.

\noindent
\textbf{Evaluation Dataset}. 
The \textit{BFCL} evaluates the LLMs ability to invoke functions in the real-world by actually triggering the API call and comparing responses, provides a comprehensive dataset comprising 4k+ instances (updating), consisting of {Non-live}, {Live} (with user-contributed complex tools avoiding contamination),  {Multi-turn} subset.
Other benchmarks, namely \textit{API-Bank} \citep{li2023api}, \textit{Nexus Raven} \citep{srinivasan2023nexusraven}, \textit{Tool-Alpaca} \citep{tang2023toolalpaca}, and \textit{Seal-Tools} \citep{wu2024seal}, are elaborated in Appendix \ref{sec:benchmark}. 

\noindent
\textbf{Baselines}
(1)\textit{ Vanilla Model}: the original model without additional training (e.g., Llama3.1-series, Qwen2.5-series).
(2) \textit{SFT Model}s: instruct models fine-tuned on supervised data, to assess whether GRPO training outperforms standard SFT, including ToolACE-8B (trained in ToolACE), xLAM-series (trained in xLAM)\citep{zhang2024xlam}, and Hammer-series (trained on xLAM with function mask to enhance generalization) \citep{lin2024hammer}.
(3) \textit{API-based} closed-source models (e.g., GPT-series, Gemini-series).
(4) \textit{R1-like Model}: models trained using GRPO with SFT as the RL paradigm, such as QwQ-32B \citep{qwq32b}, Tool-N1 series (single turn tool-use models trained in mixed ToolACE and xLAM data) \citep{zhang2025nemotron}, and ToolRL(trained in subset of mixed ToolACE and xLAM data) \citep{qian2025toolrl}.
Instead, our Tool-Zero series trained with pure RL without SFT (i.e., R1-Zero).

\begin{table*}[th]
    \small
    \centering
    \begin{tabular}{l|ccccc|c}
    \toprule
    Models & BFCL-v3 & API-Bank & SealTool & Tool-Alpaca & Nexus Raven & Avg. \\ \midrule
    \ding{168} Qwen2.5-32B-Instruct  
        & \ding{174}59.67 & \ding{174}75.87 & \ding{173}93.08 &\ding{172}65.16& \ding{173}89.12   & \ding{174}76.58 \\ 
    \ding{169} Hammer2.1-7b  
        & \ding{173}61.83 & \ding{172}81.45 & \ding{172}94.94 & \ding{173}64.60 & \ding{174}84.35   & \ding{173}77.43 \\ 
    \ding{169} ToolACE-8B  
        & \ding{175}58.42 & \ding{176}69.49 & \ding{175}89.71 & \ding{175}62.07 & \ding{175}80.23  & \ding{175}71.98 \\ 
    \ding{169} xLAM-7B-fc
        & \ding{176}54.75 & \ding{175}72.45 & \ding{176}76.90 & \ding{176}59.00 & \ding{176}57.50  & \ding{176}64.12 \\ 
    \ding{170} GPT-4o-2024-11-20
        & \ding{172}72.08 & \ding{173}80.52 & \ding{174}90.63 & \ding{174}62.37 & \ding{172}90.19  & \ding{172}79.16 \\ 
    \midrule 
    
    \ding{171} DeepSeek-R1  
        & \ding{175}56.89 & \ding{174}71.22 & \ding{175}89.97 & \ding{174}65.75 & \ding{173}82.88  & \ding{174}73.34 \\ 
    \ding{171} QwQ-32B 
        & \ding{176}53.93 & \ding{175}70.29 & \ding{174}92.94 & \ding{175}62.29 & \ding{174}63.61  & \ding{175}68.61 \\ 
    \ding{171} ToolRL-7B  
        & \ding{174}58.38 & \ding{176}67.56 & \ding{176}85.81 & \ding{172}74.13 & \ding{174}76.34   & \ding{176}68.44 \\ 
    \rowcolor{green!20} \ding{171} Tool-Zero-7B  
        & \ding{173}65.22 & \ding{173}79.85 & \ding{173}94.73 & \ding{175}65.71 & \ding{174}82.74   & \ding{173}77.65 \\ 
    \rowcolor{green!20} \ding{171} Tool-Zero-32B  
        & \ding{172}67.12 & \ding{172}81.63 & \ding{172}95.16 & \ding{173}67.38 & \ding{172}85.33   & \ding{172}79.32 \\ 
    \bottomrule
    \end{tabular}
    \caption{Comparison on more benchmarks. 
    Rankings within each column are shown with circled numbers, where smaller numbers indicate larger values.
    Tool-Zero demonstrate better performance across multiple benchmarks consistently.
    }
    \label{tab:benchmarks}
\end{table*}

\subsection{Overall Performance}
\textbf{Results on BFCL}. 
Table \ref{tab:overall_res} shows the evaluation results, covering three subset metrics. 
We observe that SFT models like ToolACE-8B and xLAM-7b-r perform well on Live (data with the same distribution as training data) due to domain-specific training but exhibit poor generalization in out-of-distribution metrics (e.g., Multi-Turn). 
In contrast, Tool-Zero series models outperform others across all metrics. For instance, Tool-Zero-7B achieves 13.32 and 19.32 improvement in Live and Multi-Turn, respectively, compared to Qwen2.5-7B-Instruct.

Pure RL paradigms outperform the SFT-then-RL approach. 
For instance, among R1-like models, Tool-Zero-7B surpasses DS-R1 by +8.33 and ToolRL-7B by +6.84.
This indicates RL better elicits intrinsic reasoning abilities from \textit{Zero} model, whereas SFT merely focuses on mimicking superficial patterns.
Notably, compared to SFT models, models trained with GRPO (Tool-N1, ToolRL, Tool-Zero) perform comparably on Live and better on Non-live and multi-turn tasks.
These results confirm that the RL paradigm is more effective for enhancing tool-integrated reasoning.

\textbf{Results on More Benchmarks}. 
Table \ref{tab:benchmarks} presents the results. 
Across different benchmarks, SFT models show inconsistent performance, while GPT-4o performs best. 
Notably, Hammer2.1-7b exhibits relatively consistent performance, attributed to its function masking techniques.
Compared to SFT and R1-like models, Tool-Zero models demonstrate significantly more stable performance, highlighting the robustness of GG-GRPO.
These findings indicate that our method generalizes effectively across various tool-use scenarios, offering new avenues for enhancing the tool-integrated reasoning capabilities of LLMs.

\subsection{Experimental Analysis}
\subsubsection{Ablation Study}
\begin{figure}[t]
  \centering
  \includegraphics[width=0.95\linewidth]{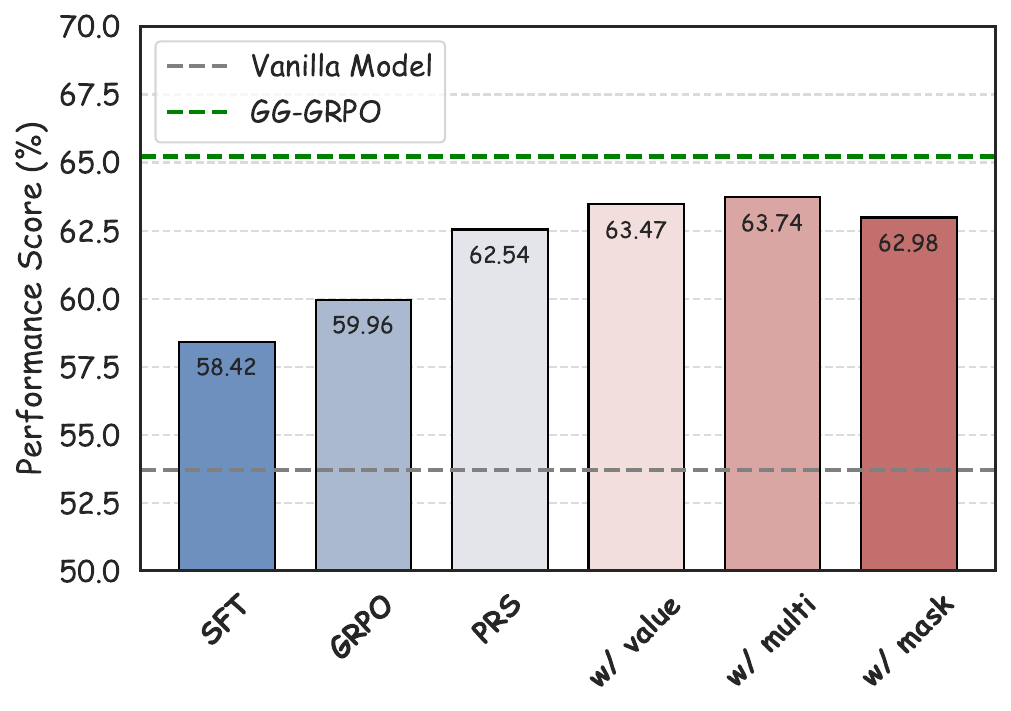}
  \caption{ Ablation study results for GG-GRPO on BFCL benchmark overall performance. }
  \label{fig:ablation_1}
\end{figure}

We conduct an ablation study for GG-GRPO, which comprises the progressive reward strategy (PRS), two tool-specific signals (multi-tool, value error), and the tool mask.
Using the Vanilla model Qwen2.5-7B-inst, we compare model training via SFT and pure GRPO training with same training data ToolACE. 
The results are presented in Figure \ref{fig:ablation_1}.
We observe that GG-GRPO achieved a +5.26 improvement compared to GRPO, and a +6.8 mprovement compared to SFT.
Experimental results demonstrate that all components contribute significantly to model performance. 
Among them, multi-tool and value error signals yield more substantial improvements compared to call pattern signals and the tool mask.

\begin{figure}[t]
  \centering
  \includegraphics[width=0.8\linewidth]{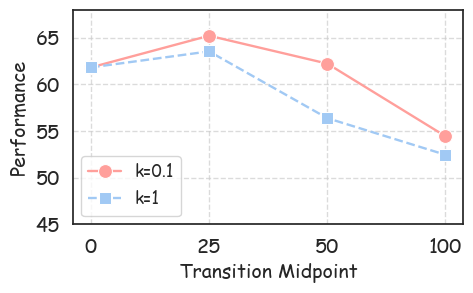}
  \caption{ Hyperparameter analysis for progressive reward strategy on BFCL benchmark overall performance. }
  \label{fig:ablation_2}
\end{figure}
Additionally, we conduct a hyperparameter ablation study on the \textit{progressive reward strategy} by varying two key parameters: transition midpoint \( t_m \in \{0, 25, 50, 100\} \) and steepness factor \( \kappa \in \{0.2, 1.0\} \) (controlling reward transition slope). 
Results (Figure \ref{fig:ablation_2}) show that a smaller transition midpoint (\( t_m = 25 \)) yields the best performance, while larger values (\( t_m \geq 50 \)) lead to degradation. 
This aligns with prior observations \citep{pan2024feedback,zhang2025nemotron} that excessive exploration in fine-grained schemes may induce reward hacking and overfitting to superficial cues. 
Also, a lower steepness factor consistently outperforms, indicating that gradual reward shaping stabilizes training. 
These findings validate the design choices in GG-GRPO's progressive reward mechanism.

\begin{table}[]
    \centering
    \small
    \begin{tabular}{lcccc}
        \toprule
        Models & Non-live & Live & Multi-turn \\ \midrule
        \multicolumn{3}{l}{\textit{In ToolACE}} & \\ \midrule
        \hspace{0.cm} w/ qwen2.5-7b & \underline{88.98} & \textbf{80.76} & \textbf{25.93} \\
        \hspace{0.cm} w/ qwen2.5-32b & {90.76} & {82.43} & {28.18} \\
        \hspace{0.cm} w/ qwen2.5-7b-inst & \textbf{89.39}  & 79.39  & 21.74 \\
        \hspace{0.cm} w/ qwen2.5-7b-coder & 88.94  & \underline{80.12}  & \underline{24.38} \\
        \midrule
        \multicolumn{2}{l}{\textit{In xLAM}} & \\ \midrule
        \hspace{0.cm} w/ qwen2.5-7b & \textbf{87.15} & \textbf{76.93} & \textbf{32.38} \\
        \hspace{0.4cm} w/o MT-Aug. & \underline{87.04}  & 74.79  & 16.18 \\
        \hspace{0.cm} w/ qwen2.5-7b-inst & 85.28  & \underline{75.32}  & \underline{29.47} \\
        \bottomrule
    \end{tabular}
    \caption{The result of data \& backbones generalizability analysis, \texttt{MT-Aug.} typos multi turn augment in Section \ref{sec:data}. }
    \label{fig:ablation_3}
\end{table}

\subsubsection{Training Data \& Backbones Generalizability}
To further validate the effectiveness of the proposed methods, we investigated the performance of our GG-GRPO across different datasets and backbone language models. 
As shown in Table \ref{fig:ablation_3}, the experimental results demonstrate that training with the Base model consistently yields better performance across various training datasets compared to the Instruct model. 
This indicates that models with stronger instruction-following capabilities do not necessarily bring greater training benefits to tool-augmented models in RL. 
We attribute this to the Base model's higher plasticity, which more easily elicits intrinsic reasoning abilities.
Additionally, when trained on different xLAM datasets, it also achieves consistently strong performance. 
Furthermore, through ablation experiments on Multi-Turn Augment in xLAM, we observed a significant increase in results from 16.18 to 32.28, highlighting the effectiveness of this augmentation strategy.
A additional study of model scale is detailed in Appendix \ref{sec:model_scale}.


\section{Conclusion}
This study firstly extends a pure rule-based RL paradigm in tool-augmented models.
Designing a dynamic generalization-guided reward to tackle the generalization limitations. 
By fostering intrinsic reasoning through progressive exploration-exploitation strategies, our approach reduces reliance on task-specific data and enhances tool-agnostic adaptability. 
Across diverse benchmarks, Tool-Zero models outperform SFT and RL-with-SFT baselines.
These results validate RL’s potential for scalable, autonomous tool learning in LLMs, advancing versatile AI agents for open-domain tasks.

\section{Limitation}
While our study has achieved notable advancements, it is important to acknowledge several limitations that could be addressed in future work.
(1) The progressive reward-switching strategy, though effective for generalization, introduces additional computational costs during the RL training phase, particularly for large models (e.g., Tool-Zero-32B). This limits scalability on resource-constrained hardware without further optimization.
(2) We acknowledge that evaluation datasets (e.g., BFCL and API-Bank) have known limitations (e.g., lengthy calling chains) arising from design preferences.

\section*{Acknowledgements}
The research in this article is supported by the New Generation Artificial Intelligence of China (2024YFE0203700), National Natural Science Foundation of China under Grants U22B2059 and 62176079.

\bibliography{custom}
\appendix

\section{Details of Experimental Setup }
\label{sec:exp_setup}

\subsection{The Implementation Settings}
The experiments were executed using the publicly accessible training framework MindSpeed-RL\footnote{https://gitee.com/ascend/MindSpeed-RL}, an end-to-end reinforcement learning acceleration framework based on the Ascend ecosystem. 
In baselines, we utilized the QwQ-32B model and DeepSeek-R1 with the thinking inference mode enabled via API calls provided by Alibaba Cloud.
Key hyperparameters included:
$\kappa = 0.1$, temperature parameter for exploration-exploitation trade-off;
Transition midpoint set to 25 (defining the inflection point in reward function scheduling).

For all the tool calls in the dataset, we use a hybrid format combining JSON structure and Python code snippets was adopted to encode various tool call format.
For the GG-GRPO (a variant of GRPO) training, model training can be done within 28 hours with 5*8 Ascend 910b NPUs per run with the following hyperparameters:

\begin{table}[ht]
    \centering
    \begin{tabular}{lc}
        \toprule  
        Category & Hyperparameter \\
        \midrule  
        \multicolumn{2}{l}{\textbf{Data Configuration}} \\
        \midrule
        Global Batch Size & 128 \\
        
        Max Prompt Length & 2048 \\
        Max Response Length & 2048 \\
        \midrule
        \multicolumn{2}{l}{\textbf{Optimization}} \\ \midrule
        Learning Rate & 5e-7 \\
        LR Decay Style & cosine \\
        Mini Batch Size & 1024 \\
        Tensor Model Parallel Size & 4 \\
        KL Loss Used & False \\  
        $\epsilon$ & 0.2 \\
        \midrule
        \multicolumn{2}{l}{\textbf{Rollout Configuration}} \\  \midrule
        Rollout Name & vllm \\
        GPU Memory Utilization & 0.9 \\
        Number of Rollouts & 8 \\ 
        Temperature & 0.8 \\  
        \midrule
        \bottomrule  
    \end{tabular}
    \caption{The configurations for RL training with GG-GRPO.}
    \label{tab:grpo_config}
\end{table}

\begin{figure*}[th]
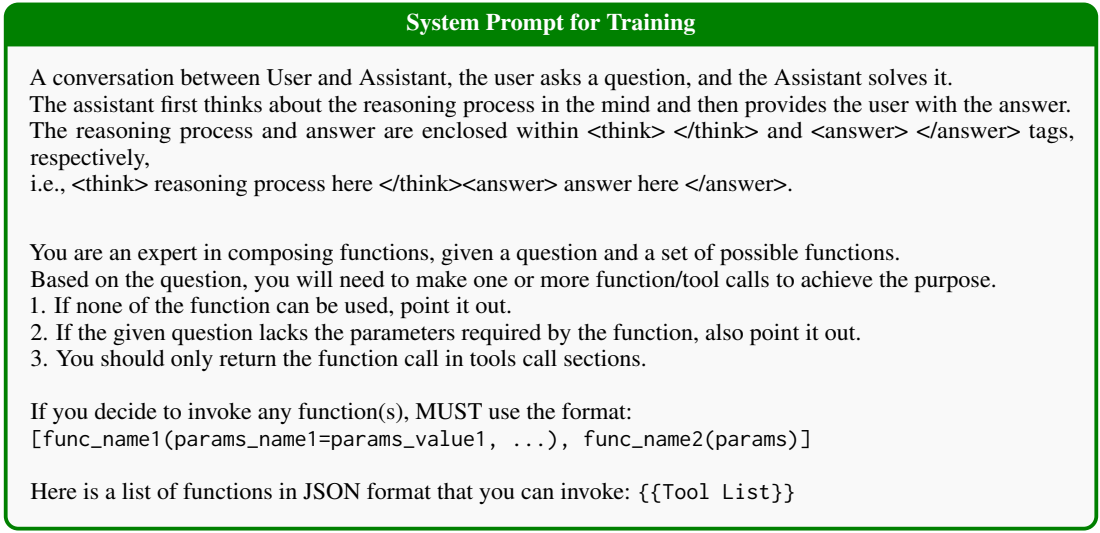

    \centering
    \begin{tcolorbox}[
        colback=gray!5!white,
        colframe=green!50!black,
        title={\bfseries System Prompt for Training}, 
        fonttitle=\bfseries\small, 
        width=0.9\textwidth, 
        left=2mm, right=2mm, 
        top=2mm, bottom=2mm, 
        center title,
        arc=4pt, 
    ]
    \small
    \noindent
    A conversation between User and Assistant, the user asks a question, and the Assistant solves it. \\
    The assistant first thinks about the reasoning process in the mind and then provides the user with the answer. \\
    The reasoning process and answer are enclosed within <think> </think> and <answer> </answer> tags, respectively, \\
    i.e., <think> reasoning process here </think><answer> answer here </answer>. \\[1.5ex] 
    
    You are an expert in composing functions, given a question and a set of possible functions. \\
    Based on the question, you will need to make one or more function/tool calls to achieve the purpose. \\
    1. If none of the function can be used, point it out.\\
    2. If the given question lacks the parameters required by the function, also point it out.\\
    3. You should only return the function call in tools call sections.\\
    
    If you decide to invoke any function(s), MUST use the format: \\
    \texttt{[func\_name1(params\_name1=params\_value1, ...), func\_name2(params)]} \\
    
    Here is a list of functions in JSON format that you can invoke: \texttt{\{\{Tool List\}\}}
    \end{tcolorbox}
    \caption{The system think prompt with Python code format for RL Training. The prompt guides the LLM to explicitly separate reasoning process and answer. }
    \label{train_prompt}
\end{figure*}

\section{Benchmark \& Metric Details.}
\label{sec:benchmark}
The BFCL is an evolving benchmark. For our study, we utilized the version checked out on February 26, 2024. 
Other benchmarks include: (1) API-Bank with 314 tool-use dialogues and 753 API calls, evaluating known API invocation (L-1) and candidate list retrieval/calling (L-2), we report their average result in evaluation; (2) Nexus Raven API Evaluation offering 318 test examples across 65 APIs for function-calling assessment; (3) Tool-Alpaca’s 271 synthetic tool-use instances in 50 categories (100 simulated tests used); (4) Seal-Tools, a recent benchmark with 4,076 auto-generated APIs across life domains.
The BFCL assesses models using Abstract Syntax Tree Evaluation and Executable Function Evaluation Accuracy, and the other benchmarks assesses models using Function and Parameter matching F1 score \citep{lin2024hammer}.
For the evaluation, each model was administered the same system prompt as its training counterpart.

\textbf{Dynamic Tool Use}. Moreover, we conducted a comprehensive analysis to verify the dynamic nature of tool selection in our approach.
It evaluates whether the candidate tools utilized across different datasets exhibit sufficient variability to necessitate a dynamic retrieval approach, thereby justifying our method's design choices related to scalability and adaptability to unseen tools.

Our analytical procedure consisted of two main phases:
First, we examined the distribution of candidate tools across various benchmark datasets by quantifying the total number of unique tools in each dataset. This provided a baseline understanding of the scale and variability in toolset sizes, as shown in Table~\ref{tab:tool_counts}.
Table \ref{tab:tool_counts} presents the number of unique tools identified in each dataset, revealing substantial variation in toolset sizes across different benchmarks, ranging from 41 tools in Tool-Alpaca to 25,771 tools in ToolACE.

Second, to measure the degree of overlap between toolsets of different datasets, we calculated the overlap rate using the formula:
\begin{equation}
\text{overlap rate} = \frac{|A \cap B|}{\min(|A|, |B|)} \times 100\%
\end{equation}
where \(A\) and \(B\) represent the sets of tools from two different datasets, \(|A \cap B|\) denotes the cardinality of their intersection, and \(\min(|A|, |B|)\) represents the size of the smaller set.
Table \ref{tab:overlap_rates} summarizes the overlap rates between the toolsets of different datasets. The results demonstrate minimal to non-existent overlap, with rates ranging from 0\% (e.g., between xLAM and API-Bank) to a maximum of 47.1\% (between ToolACE and SealTool).

\begin{table}[th]
  \centering
  \begin{tabular}{@{}lc@{}}
    \toprule
    Dataset      & Number of Tools \\ \midrule
    BFCL-v3      & 2031            \\
    API-Bank     & 50              \\
    SealTool     & 1084            \\
    Tool-Alpaca  & 41              \\
    Nexus Raven  & 65              \\
    ToolACE      & 25771           \\
    xLAM         & 3605            \\ \bottomrule
  \end{tabular}
  \caption{Number of Unique Tools in Each Dataset}
  \label{tab:tool_counts}
\end{table}

\begin{table*}[th]
  \centering
  \begin{tabular}{@{}lccccc@{}}
    \toprule
    Train/Eval & BFCL-v3 & API-Bank & SealTool & Tool-Alpaca & Nexus Raven \\ \midrule
    ToolACE    & 5.5     & 20.0     & 47.1     & 2.4         & 15.3        \\
    xLAM       & 1.8     & 0.0      & 0.0      & 2.4         & 0.0         \\ \bottomrule
  \end{tabular}
  \caption{Toolset Overlap Rates Between Datasets (\%)}
  \label{tab:overlap_rates}
\end{table*}

\begin{table*}[th]
    \small
    \centering

    \begin{tabular}{lc|ccccc|c}
        \toprule
        \textbf{BackBone} & \textbf{Algorithm} & \textbf{BFLC-v3} & \textbf{API-Bank} & \textbf{SealTool} & \textbf{Tool-Alpaca} & \textbf{Nexus Raven} & \textbf{Avg.} \\
        \midrule
        Qwen2.5-7B-Instruct & SFT     & 38.95 & 32.98 & 56.44 & 43.18 & 52.13 & 44.73 \\
        Qwen2.5-7B & GG-PPO     & 55.34 & 69.93 & 82.75 & 59.24 & 78.13 & 69.07 \\
        Qwen2.5-7B & GRPO & 59.96 & 74.27 & 91.71 & 59.71 & 76.28 & 72.38 \\
        \midrule
        Qwen2.5-1.5B & GG-GRPO       & 50.39 & 68.65 & 89.85 & 47.28 & 62.46 & 63.73 \\
        Qwen2.5-3B   & GG-GRPO       & 57.29 & 74.25 & 93.64 & 57.32 & 71.85 & 70.87 \\
        Qwen2.5-7B   & GG-GRPO       & 65.22 & 79.85 & 94.73 & 63.71 & 82.74 & 77.32 \\
        Qwen2.5-32B  & GG-GRPO       & 67.12 & 81.63 & 95.16 & 64.38 & 85.33 & 78.99 \\
        \bottomrule
    \end{tabular}
    \caption{Performance comparison across training methods and model scales.}
    \label{tab:performance_comparison}
\end{table*}

These findings confirm that candidate toolsets exhibit high variability across different datasets, with no fixed or universal toolset configuration. The low overlap rates indicate that tools relevant to one dataset are often irrelevant to others, supporting the need for a dynamic tool retrieval mechanism. Consequently, our model's approach of learning generalizable tool-usage patterns rather than memorizing specific tools enables effective adaptation to unseen tool-use scenarios, addressing the scalability concerns inherent in fixed-toolset approaches.

\section{System Thinking Template}
We adopt a lightweight prompting schema to elicit tool-use capabilities from the LLM, drawing inspiration from prior work \cite{deepseekai2025r1,openr1}. 
As illustrated in Figure \ref{train_prompt}, the template explicitly instructs the model to encapsulate intermediate reasoning within \texttt{<think>...</think>} tags, followed by the final answer enclosed in \texttt{<answer>...</answer>} tags. 
By allowing the model greater freedom in articulating its reasoning process, we aim to enhance generalization across diverse tool integration scenarios. Additionally, this design facilitates seamless adaptation to complex tool-augmented reasoning tasks.

\section{Model Scale Analysis}
\label{sec:model_scale}
To address concerns about the model scalability of the proposed method, the experiment on scalability was conducted. 
the objective was to analyze our method's performance across different-scale base models. 
The operation involved evaluating the performance of Tool-Zero models with various base model scales (1.5B, 3B, 7B, 32B) across datasets like \textit{BFCL-v3, API-Bank}, etc. 
The results, in Table~\ref{tab:performance_comparison}, showed that the method performed strongly on 7B and 3B base models, while results on 1.5B and 32B models were relatively lower, indicating partial scalability and room for optimization in extreme-scale scenarios.

Additionally, for the experiment on algorithmic effectiveness, the objective was to evaluate the reward function on native PPO \citep{song2025r1,zeng2025simplerl} (i.e., GG-PPO) and GRPO algorithms (i.e., GG-GRPO), as shown in Table~\ref{tab:performance_comparison}.
The results revealed that our reward module with GRPO outperformed PPO, demonstrating robust and consistent gains. 
This highlights GRPO's superior adaptability to the reward framework for tool learning, which can be attributed to the reward function being designed for rule-based rewards in algorithms like GRPO (unlike PPO, which is natively for model-based rewards) and recent studies showing GRPO's advantage in achieving "aha moments" compared to PPO-family algorithms \citep{qian2025toolrl,deepseekai2025r1,muennighoff2025s1}.

\end{document}